\documentclass[11pt,letterpaper]{article}
\usepackage[letterpaper]{geometry}
\usepackage{acl2012}
\usepackage{times}
\usepackage{latexsym}
\usepackage{amsmath}
\usepackage{multirow}
\usepackage{url}

\usepackage{url}
\usepackage{graphicx}
 \usepackage{subfigure}
\graphicspath{ {Figure/} }
\usepackage{amsmath, amsfonts}
\usepackage{algorithm}
\usepackage[noend]{algpseudocode}
\usepackage{multirow}
\DeclareMathOperator*{\argmax}{arg\,max}
\makeatletter
\newcommand{\@BIBLABEL}{\@emptybiblabel}
\newcommand{\@emptybiblabel}[1]{}
\makeatother
\usepackage[hidelinks]{hyperref}
\title{Geometry of Compositionality}

\author{Hongyu Gong \and Suma Bhat \and Pramod Viswanath \\
hgong6@illinois.edu, spbhat2@illinois.edu, pramodv@illinois.edu
\\
Department of Electrical and Computer Engineering\\ 
University of Illinois at Urbana Champaign,USA}
\date{}

\begin{document}
\maketitle
\begin{abstract}
This paper proposes a simple test for compositionality (i.e., literal usage) of a word or phrase in a {\em context-specific} way. The test is computationally simple,  relying on no external resources and only uses a set of trained word vectors.  Experiments show that the proposed method is competitive with state of the art and displays high accuracy in context-specific compositionality detection of a variety of natural language phenomena (idiomaticity, sarcasm, metaphor) for different datasets in multiple languages. The key insight is to connect compositionality to a curious geometric property of word embeddings, which  is of independent interest.
\end{abstract}

\section{Introduction}
Idiomatic expressions and figurative speech are  key components of the creative process that embodies natural language. One expression type is  
multiword expressions (MWEs) --   phrases with semantic idiosyncrasies that cross word boundaries \cite{sag2002multiword}. % The difference of strong MWEs from weak MWEs (or called compounding phrases) is that the meaning of a strong MWE cannot be determined simply from its component words. For example, the semantic meaning of the compounding phrase \emph{red sweater} is a composition of meanings of component words \emph{red} and \emph{sweater}. However, the meaning of the MWE \emph{of course} has nothing to do with the semantics of words \emph{of} and \emph{course}. Strong MWEs instead of single words act as a basic unit in some linguistic applications like machine translation \cite{ren2009improving}. 
 Examples of MWEs include \textit{by and large}, \textit{spill the beans} and \textit{part of speech}. As such, these phrases  are  idiomatic, in that their meanings cannot be inferred from the meaning of their component words, and are hence termed \textit{non-compositional} phrases % or {\em strong MWEs} 
 as opposed to being \textit{ compositional} phrases. % or {\em weak MWEs}. 

%The components of an idiomatic MWE have degrees of semantic compositionality with varying constraints on word order and composition \cite{nunberg1994idioms}, e.g. {\em in short} and *{\em in very short}, whereas {\em spill the beans} and {\em spilled the beans}.
A particularly intriguing aspect of MWEs is their ability to take on degrees of compositionality depending on the context they are in. For example, consider two contexts in which the phrase \textit{bad egg} occurs. \\
(1) Ensure that one \textbf{bad egg} doesn't spoil good businesses for those that care for their clientele.\\
(2) I don't know which hen is laying the \textbf{bad egg} but when I crack it, it explodes!  It is all creamy yellowish with very little odor. \\
In (1), the phrase has a non-compositional interpretation to mean `an unpleasant person', whereas in (2), the phrase has the meaning of a noun phrase whose head is {\em egg}  and modifier is {\em bad}. 
This context-dependent degree of compositionality of an MWE poses significant challenges to natural language processing applications. In machine translation, instead of processing the MWE as a whole,  literal translation of its components  could result in a  meaningless phrase in the target language, e.g.,  {\em chemin de fer} from French to English to be \textit{way of iron} in place of \textit{railway} \cite{bouamor2012identifying}. In information retrieval, the retrieved document matching a component word is irrelevant given the meaning of the MWE {\em hot dog}. %\cite{evans1996noun}
 Hence, identifying the compositionality of  MWEs is an important subtask in all these systems. 

As another example, consider the word {\em love} in the following two contexts. In the first: ``I {\bf love} going to the dentist. Been waiting for it  all week!'', the word has a non-literal (hence non-compositional) and \textit{sarcastic} interpretation to actually mean the exact opposite of the literal (compositional) sense, which is to ``like". In the second: 
 ``I  {\bf love} strawberry ice cream;  it's simply my favorite",  
 the same word has the compositional meaning.  Again, the degree of compositionality is crucially context-dependent. 
 
Yet another example of compositionality involves {\em metaphors}. Consider the word {\em angel} in the following two contexts: \\
(1) The girl is an \textbf{angel}; she is helpful to the children. \\
(2) The \textbf{angels} are sure keeping busy, what with all his distractions and mishaps. \\
In (1) the word has a figurative sense (i.e., non-compositional  interpretation) whereas in (2), the word has the compositional meaning of  a ``divine being".  Again, the degree of compositionality is crucially context-dependent.  %This same phenomenon is true of metaphors: the word {\em ship} in the sentence ``Camel is the {\em ship} of the desert" is a noncompositional usage while the usage in ``The US navy is building the ship of the future'' is not. 

In this paper our focus is to  decide the compositionality of a word or a phrase using its {\em local linguistic context}. 
Our approach only relies on the use of word embeddings, which capture the ``meaning" of a word using a low-dimensional vector. Our compositionality prediction algorithm brings two key innovations: (1) It  leverages the crucial contextual information that dictates the compositionality of a phrase or word; (2) The prediction mechanism is completely independent of external linguistic resources. 
Both these are significant improvements over recent works with similar goals: compositionality of MWEs \cite{salehi2015word},  works on sarcasm \cite{wallace2014humans} and metaphor detection \cite{tsvetkov2014metaphor} (the latter works rely significantly on external linguistic resources and access to labeled training data). %We rely on the geometry of the phrase vector in the space spanned by the word vectors of its context to predict the degree of compositionality. 
% * <hgong6@illinois.edu> 2016-11-22T14:52:51.801Z:
%
% ^.
To the best of our knowledge, this is the first study on {\em context-dependent} phrase compositionality in combination with word embeddings and the first resource-independent study on sarcasm and metaphor identification. This work is centered around two primary questions:\\
(1) How can the semantics of a long context be represented by word embeddings?\\
(2) How can we decide the compositionality of a phrase based on its embeddings and that of its context? 
    %\item Whether can the method of compositionality detection be generally applied to different languages?

We answer these questions by connecting the notion of compositionality to a geometric property of word embeddings. The key insight is that the context word vectors (suitably compressed) reside roughly in a low dimensional {\em linear} subspace and compositionality turns out to be related to the projection of the word/phrase embeddings (suitably compresed to a single vector) onto this context subspace. 

The key justification for our approach comes from empirical results that outperform state of the art methods  on many metrics, while being competitive on the others.  
We use three standard datasets spanning two MWE construction types (noun compounds and verb particle constructions) in two languages (English and German) in addition to a dataset in Chinese (heretofore unexplored language), and standard datasets for detection of metaphor and sarcasm in addition to a new dataset for sarcasm detection from Twitter.  We summarize our contributions:\\
\textbf{Compositional Geometry}: We show that a word (or MWE) and its context are geometrically related as jointly lying in a {\em linear} subspace, when it appears in a compositional sense, but not otherwise.\\ 
\textbf{Compositionality decision}: The {\em only} input to the algorithm is a set of trained word vectors after the preprocessing step of removing function words on which,  the algorithm performs a simple principle component analysis (PCA)  operation.\\
\textbf{Multi-lingual applicability}: The algorithm is very general, relies on no external resources  and is agnostic to the specifics of one language; we demonstrate strong test results across  different languages.

We begin next with a discussion of the geometry of compositionality leading directly to our context-based algorithm for compositionality detection. The test is competitive with or superior to state of the art in a variety of contexts and languages, and in various metrics.

\begin{figure}[htbp]
\centering
\includegraphics[width=0.45\textwidth]{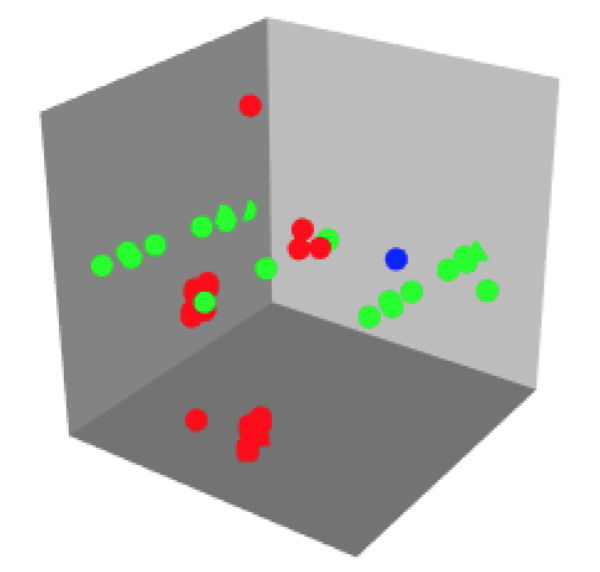}
%\captionsetup{justification=centering}
\caption{\quad Geometry of phrase and context. \protect\\ The compositional context embeddings of \emph{cutting edge} are denoted by green points, and the non-compositional context embeddings  by red points. The  embedding of phrase \emph{cutting edge} is denoted by the blue point. Note that the  phrase embedding is very close to the space of the compositional context while being farther from the space of its non-compositional context.}
\label{fig:geom}
\end{figure}

%-------------------   section 2 after revision -----------------------

\section{Compositionality and the Geometry of Word Embeddings }
\label{sec:model}
Our goal is to detect the compositionality level of a given occurrence of a {\em word/phrase}  within a {\em sentence} (the context).  Our {\bf main contribution} is the discovery of a geometric property of vector embeddings of context words (excluding the function words)   within a sentence: they roughly occupy  a low dimensional linear subspace which can be empirically extracted via a standard dimensionality reduction technique:  principal component analysis (PCA) \cite{Shlens14}.

We stack the $d$-dimension vectors $v_{1},\ldots , v_{n}$, corresponding to $n$ words in a sentence, to form a $d \times n$ matrix  $X$. PCA finds a $d \times m  (m < n)$ matrix $X'$ which maximizes the data variance with reduced dimension. Here $X'$ consists of $m$ new vectors, $v_{1}', ..., v_{m}'$. Now the original data $X$ is represented by fewer vectors of $X'$, where  vectors $v$ and $v'$ are $d$-dimensional ($m$ is chosen such that a large enough fraction -- a hyperparameter -- of the variance of $X$ is captured in $X'$). %PCA is a standard technique commonly used for data dimension reduction. 

When the phrase of interest occurs in a compositional sense, then the phrase's compositional embedding is roughly close to the subspace associated with the context embeddings (extracted using PCA from context words). Intuitively this happens because  compositionality  is tantamount to individual words themselves being directly related (i.e., occur together often enough) to (a majority  of) the context words. 

We illustrate this phenomenon via an example found in Table~\ref{tab:mweAndCxt}. Consider the phrase ``cutting edge". When words like \emph{sharp, side} and \emph{tool} appear in the context, ``cutting edge'' tends to have a  compositional meaning. Conversely, when words like \emph{productions, technology} and \emph{competitive} are in the context, ``cutting edge'' is more likely to be an idiom.  We project the embeddings of the  phrase and the two contexts to three-dimensions to visualize the geometric relationship, cf.\ Figure \ref{fig:geom}. 

It is immediate that the  phrase embedding occupies the same subspace as the  context when it is used in the compositional sense, while it is far from the subspace of the context when used in the non-compositional sense. 
The precise formulation of the projection operations used in this illustration is discussed next. 
 
% This subsection is to replace both Compositional Phrase 
%\subsection{Compositional Semantic Representation}
Suppose a sentence $t$ consists of $n$ content words $\{w_{1}, ..., w_{n}\}$ with respective vector embeddings $\{v_{1}, ..., v_{n}\}$. Two possible representations of the ``meaning'' of  $t$ are the following: \\
\textbf{\em average vector representation}: $v_{t} = v_{1} + ... + v_{n}$, adding all component word vectors together, as in \cite{mitchell2010composition} and several works on phrase2vec \cite{gershman2015phrase} and sentence2vec \cite{faruqui2015retrofitting,wieting2015towards,DBLP:journals/corr/AdiKBLG16}. \\
%(related works detailed  in the supplement). \\
\textbf{\em PCA subspace representation}: Denote the word vectors by  $X = [v_{1}, \ldots, v_{n}]$, and the PCA output $X' = [v_{1}', ..., v_{m}']$, where $v_{i}'$ are the principal components extracted from X using the PCA operation. Now the sentence $t$ is represented by the (span of columns of) matrix $X'$ instead of a single vector as in average vector representation. Choosing to {\em represent the sentence by multiple vectors is a key innovation of this paper} and is fairly critical to the empirical results we demonstrate. 

\renewcommand{\arraystretch}{1.3}
\begin{table*}[!htbp]
\centering
\begin{tabular}{|c|c|c|}
\hline
\multicolumn{1}{|c|}{Phrase} & \multicolumn{1}{c|}{Compositional Context} & \multicolumn{1}{c|}{Non-compositional Context} \\ \hline
{\bf cutting edge} & \parbox[c]{0.4\textwidth}{the flat part of a tool or weapon that (usually) has a {\bf cutting edge}. Edge - a sharp side.} & \parbox[c]{0.4\textwidth}{while creating successful film and TV productions, a {\bf cutting edge} artworks collection.} \\ \hline
\textbf{ground floor} & \parbox[c]{0.4\textwidth}{Bedroom one with en-suite is on the \textbf{ground floor} and has a TV. Furnished with king size bed, two bedside chests of drawers with lamps.} & \parbox[c]{0.4\textwidth}{Enter a business organization at the lowest level or from the \textbf{ground floor} or to be in a project undertaking from its inception}. \\ \hline
\textbf{free lunch} & \parbox[c]{0.4\textwidth}{you will be able to get something awesome at participating pizza factory restaurants: a \textbf{free lunch} special or a scrumptious buffet on veterans day} & \parbox[c]{0.4\textwidth}{travelers on highways in the u.s. have enjoyed what felt like a \textbf{free lunch}. true, gas taxes are levied to offset the cost of constructing and maintaining roadways} \\ \hline
%\textbf{算账} & \parbox[c]{0.4\textwidth}{对特定对象的经济活动进行记账、%\textbf{算账}、报账，为各有关方面提供会计信息的功能。} & \parbox[c]{0.4\textwidth}{刘松柏因为支持宪宗二子李恽为皇帝，被穆宗秋后\textbf{算账}，一生从军塞外，不得回城。} \\ \hline
%{\bf 结果} & \parbox[c]{0.4\textwidth}{天文台其馀29天均有录得降雨量。{\bf 结果}，1966年6月整月雨量达962.9毫米，创下了当时6月份最高雨量纪录。} & \parbox[c]{0.4\textwidth}{混合芽大多数为顶芽，少有腋芽。混合芽萌发后先长枝生叶，后从顶端生出花穗，开花{\bf 结果}。} \\ \hline
\end{tabular}
%\caption{Examples of phrases in English and Chinese, whose compositionality depends on the context.}
\caption{Examples of English phrases, whose compositionality depends on the context.}
\label{tab:mweAndCxt}
\end{table*}

%\textcolor{red}{*** explanation of compositionality detection ***}

Note that  the PCA operation returns a $(d\times m)$ matrix $X'$ and thus PCA is used to reduce the ``number of word vectors" instead of the embedding dimension. In our experiments, $d=200$, $n\approx 10-20$, and $m\approx 3$. PCA extracts the most important information conveyed in the sentence with only $m$ vectors. Further, we only take the linear span of the $m$ principal directions (column span of $X'$), i.e., a {\em subspace} as the  representation of sentence $t$. 

Let $p$ be a single word (in the metaphor and sarcasm settings) or a bigram phrase (in the MWE setting) that we would like to test for compositional use. Suppose that $p$ has a single-vector representation $v_p$, and the context embedding is represented by the subspace $S_c$ spanned by the $m$ vectors $(v_{1}', \ldots , v_{m}')$. Our test involves projecting the phrase embedding $v_p$ on the context subspace $S_c$. Denote the orthogonal projection vector by $v_{p}'$, where $v_{p}'$ lies in $S_c$, and $v_{p}'=\argmax\limits_{v \in \mathbb{R}^d}\frac{v^Tv_{p}}{\lVert v\rVert \cdot \lVert v_{p}\rVert}$. 

\noindent {\bf Compositionality Score} is the cosine distance between $v_{p}$ and $v_{p}'$ (the  inner product between the vectors normalized by their lengths); this measures the degree to which the word/phrase meaning agrees with its context: {\em the larger the cosine similarity, the more the compositionality}. 

%\textcolor{red}{*** explanation of compositionality detection ***}

Based on the  commonly-used distributional hypothesis that the word or phrase meaning can be inferred from its context \cite{rubenstein1965contextual}, we note that the {\em local} context (neighboring words) is crucial in deciphering the compositional sense of the word or phrase. This is in contrast to prior works  that  use the global context alone (the whole document or corpus), without accounting for the context-dependence  of polysemy \cite{reddy2011empirical}.

At times, the  word(s) being tested  exhibit polysemous behavior (example: \emph{check} in \emph{blank check}) \cite{polysemymu}. In such cases, it makes sense to consider multiple embeddings for different word senses (we use MSSG representations \cite{neelakantan2015efficient}): each word has a single global embedding and two sense embeddings. We propose to use global word embeddings to represent the context, and use sense embeddings for phrase  semantics, allowing for multiple compositionality scores. We then measure the relevance between a phrase and its context by the {\em maximum} of the different compositionality scores.

Our compositionality detection algorithm uses only two hyperparameters: {\em variance ratio} (used to decide the amount of variance PCA should capture) and {\em threshold} (used to test if the compositionality score is above or below this value). Since compositionality testing is essentially a supervised learning task: in order to provide one of two labels, we  need to tune these parameters based on a  (gold) training set. We  see in the experiment sections that these parameters are robustly trained on small training sets and are fairly invariant in their values across different datasets, languages and tasks (variance ratio equal to about 0.6 generally achieves good performance).

\begin{table}[!hbtp]
\centering
\label{tab:twoSense}
\resizebox{0.42\textwidth}{!}{
\begin{tabular}{|c|c|c|c|c|}
\hline
 & \begin{tabular}[c]{@{}c@{}}\small{English}\\ \small{(CBOW)}\end{tabular} & \begin{tabular}[c]{@{}c@{}}\small{English}\\ \small{(MSSG)}\end{tabular} & \begin{tabular}[c]{@{}c@{}}\small{Chinese}\\ \small{(CBOW)}\end{tabular} & \begin{tabular}[c]{@{}c@{}}\small{Chinese}\\ \small{(MSSG)}\end{tabular} \\ \hline
\begin{tabular}[c]{@{}c@{}}\small{avg phrase}\\ \small{avg context}\end{tabular} & 80.3 & 82.7 & 78.1 & 50 \\ \hline
\begin{tabular}[c]{@{}c@{}}\small{pca phrase}\\ \small{avg context}\end{tabular} & 59.1 & 70.2 & 50.7 & 50.7 \\ \hline
\begin{tabular}[c]{@{}c@{}}\small{avg phrase}\\ \small{pca context}\end{tabular} & 82.7 & 84.6 & 80.5 & 75 \\ \hline
\begin{tabular}[c]{@{}c@{}}\small{pca phrase}\\ \small{pca context}\end{tabular} & 85.6 & \textbf{86.1} & 81.3 & \textbf{88.3} \\ \hline
\end{tabular}}
\caption{Accuracy values (\%) for Experiment I: Compositionality detection from contexts} %Context-sensitive MWEs in English and Chinese}
\end{table}

% ENC table
\begin{table*}[!htbp]
\centering
\label{tab:enc}
\resizebox{1.0\textwidth}{!}{
\begin{tabular}{|c|c|c|c|c|c|c|c|}
\hline
 & & \multicolumn{3}{c|}{First Component} & \multicolumn{3}{c|}{Second Component} \\ \hline
Dataset & Method & \small{Precision (\%)} & \small{Recall (\%)} & \small{F1 score (\%)} & \small{Precision (\%)} & \small{Recall (\%)} & \small{F1 score (\%)} \\ \hline
\multirow{4}{*}{\begin{tabular}[c]{@{}c@{}} ENC\\ dataset \end{tabular}} & \small{PMI} & 50 & 100 & 66.7 & 40.4 & 100 & 57.6  \\ 
\cline{2-8}
%\small{LCS} & 60.0 & 77.7 & 67.7 & 81.6 & 68.1 & 64.6  \\
%\small{DS} & 62.1 & 88.6 & 73.0 & 80.5 & 86.4 & 71.2  \\ \hline
% & \small{ALLDEFS+SYN} & 81.2 & 88.1 & \textbf{84.5} & 87.3 & 80.6 & 69.8 \\
 & \small{ITAG+SYN} & 64.5 & 90.9 & 75.5 & 61.8 & 94.4 & \textbf{74.7} \\
\cline{2-8} 
% & \small{Avg Cxt (CBOW)} & 58.9 & 97.7 & 73.5 & 52.2 & 100 & 68.6 \\
 & \small{Avg Cxt (MSSG)} & 68.5 & 79.5 & 73.7 & 61.2 & 83.3 & 70.6 \\
\cline{2-8} 
 & \small{SubSpace (CBOW)} & 78.4 & 90.9 & \textbf{84.2} & 67.44 & 80.6 & \textbf{73.44} \\ 
% & \small{CB (MSSG)} & 68.9 & 90.9 & 78.4 & 57.6 & 94.4 & 71.6 \\
\hline
\multirow{4}{*}{\begin{tabular}[c]{@{}c@{}} EVPC\\ dataset \end{tabular}} & \small{PMI} & 22.2 & 68.4 & 33.5 & 53.0 & 80.2 & 63.8  \\ 
\cline{2-8}
%\small{LCS} & 36.5 & 49.2 & 39.3 & 61.5 & 63.7 & 60.3  \\
%\small{DS} & 32.8 & 34.1 & 33.5 & 80.9 & 19.6 & 29.7  \\ \hline
% & \small{ALLDEFS+SYN} & 37.4 & 70.9 & \textbf{48.9} & 80.4 & 65.9 & 63.0 \\
 & \small{ALLDEFS} & 25.0 & 97.4 & 39.8 & 53.6 & 97.6 & \textbf{69.2} \\
\cline{2-8}
% & \small{Avg Cxt (CBOW)} & 27.1 & 92.1 & 41.9 & 58.7 & 74.4 & 65.6 \\
 & \small{Avg Cxt (MSSG)} & 33.8 & 60.5 & 43.4 & 58.0 & 80.2 & 67.3 \\
\cline{2-8}
% & \small{CB (CBOW)} & 34.04 & 84.2 & \textbf{48.5} & 53.8 & 97.6 & 69.4 \\ 
 & \small{SubSpace (MSSG)} & 31.4 & 86.8 & \textbf{46.2} & 54.4 & 100 & \textbf{70.5} \\
\hline
\multirow{3}{*}{\begin{tabular}[c]{@{}c@{}} GNC\\ dataset \end{tabular}} & \small{PMI} & 44.2 & 99.0 & 61.1 & 26.4 & 98.4 & 41.7  \\
\cline{2-8}
 & \small{Avg Cxt (CBOW)} & 45.4 & 92.6 & 60.6 & 29.0 & 95.4 & 44.4 \\
% & \small{Avg Cxt (MSSG)} & 44.0 & 77.8 & 56.2 & 31.7 & 67.7 & 43.1 \\
\cline{2-8}
% & \small{Subspace (CBOW)} & 45.8 & 95.4 & 61.8 & 32.2 & 75.4 & 45.1 \\
 & \small{SubSpace (MSSG)} & 45.5 & 99.1 & \textbf{62.4} & 30.9 & 86.2 & \textbf{45.5} \\ \hline
\end{tabular}}
\caption{Experiments on ENC, EVPC and GNC Datasets.}
\label{table:mwe-condensed}
\end{table*}

%-------------------   section 2 after revision ------------------------

\section{MWE Compositionality  Detection }
\label{sec:compo_detect}
We evaluate our context-based compositionality detection method empirically by considering 3 specific, but vastly distinct, tasks: a) Predicting the compositionality of phrases that can have either the idiomatic sense or the literal sense depending on the context (the focus of this section), b) Sarcasm detection at the level of a specific word and at the level of a sentence, and c) Detecting whether a given phrase has been used in its metaphoric sense or literal sense.  The latter two tasks are the focus of the next two sections.  For each of the tasks we use standard datasets  used in state-of-the-art studies, as well as those we specifically constructed  for the experiments.   We include datasets in German and Chinese in addition to those available in English to highlight the multi-lingual and language-agnostic capabilities of our algorithm. %, using the datasets described in Section \ref{sec:dataset}. 

The training corpus of embeddings in English, Chinese and German are obtained from polyglot \cite{polyglot:2013:ACL-CoNLL}. Two types of word embeddings are used in the experiments: global ones using CBOW of word2vec \cite{mikolov2014word2vec}, and sense-specific ones  using NP-MSSG of MSSG \cite{neelakantan2015efficient}.

\subsection{Experiment I: Phrase Compositionality} %Bi-Contexts for Phrases}
\label{exp:bicontext}
% bi-context dataset description

In this part, we evaluate the performance of our algorithm in capturing the semantics of the context and predicting the compositionality of %polysemous 
phrases, which we cast as a binary classification task -- to decide the phrase compositionality in each context. With reference to the examples in  Table~ \ref{tab:mweAndCxt}, the task is to predict that the phrase \textit{cutting edge} is used in its compositional sense in the first instance and a non-compositional one in the second. We perform experiments with different word embeddings (CBOW and MSSG), as well as different composition representations for both the phrase and the context (average and PCA).

\textbf{Bi-context Dataset}:
We construct 2 datasets \footnote{available at: \url{https://github.com/HongyuGong/Geometry-of-Compositionality.git}} (one for English  and the other for Chinese) consisting of a list of %polysemous 
 phrases and their respective contexts (compositional and non-compositional). 
The English dataset contains 104 polysemous phrases which are obtained from the idiom dictionary \cite{engDict}, and the Chinese dataset consists of 64 phrases obtained from \cite{zhDict}. Their respective contexts are extracted from the corpus provided by polyglot or electronic resources \cite{googleBook}. Native English and native Chinese speakers annotated the phrase compositionality for each context. 

\textbf{Detection Results}:
%Some examples in bi-context datasets are shown (cf. Table \ref{tab:mweAndCxt}), 
We used both average and PCA subspace representations for the target phrase and its context. The results, shown as accuracy values, obtained by comparing the predicted labels with the gold labels provided by human annotators, are available in Table \ref{tab:twoSense}. Since the average vector representation is commonly used in recent works, we take ``avg phrase + avg context'' as our baseline.
We note that having a PCA approximation for both the phrase and the context,  and the use of MSSG embedding yielded the best accuracy  for this task in both the English and the Chinese datasets; this is an instance where the PCA subspace representation is superior to the average representation.   %The reason that PCA subspace representation outperforms average approximation is that PCA approximation is more robust to noise in the context. Even when there are only a few contextual words related with the phrase of interest, average sentence approximation tends to decide that the phrase is literal. However, PCA approximation would filter the distracting noise, and capture the main information of the context. 
We believe that unsupervised improvement beyond the fairly high accuracy rates is likely to require substantially new ideas as compared to those in this paper. % and prior work. 

\subsection{Experiment II: Lexical Idiomaticity}
\label{exp:component}
Unlike compositionality detection in Experiment I, here we detect component-wise idiomaticity of a two-word phrase in this experiment. For example, ``spelling'' is literal while ``bee'' is idiomatic in the phrase ``spelling bee''.  %{\tt How about an example to say what we mean by component-wise idiomaticity?}  
Modifying our method slightly, we take the cosine distance between the embedding of the target word (the first or the second word) and its projection to the space of its context as the measurement of lexical idiomaticity. The smaller the cosine distance, the more idiomatic the component word is. Here we use three datasets available from prior studies for the same task -- ENC, EVPC and GNC -- and compare our results with the state-of-art in idiomaticity detection.

% sarcasm
\begin{figure*}[htbp]
\centering
\begin{minipage}{0.32\textwidth}
\centerline{\includegraphics[width=\linewidth]{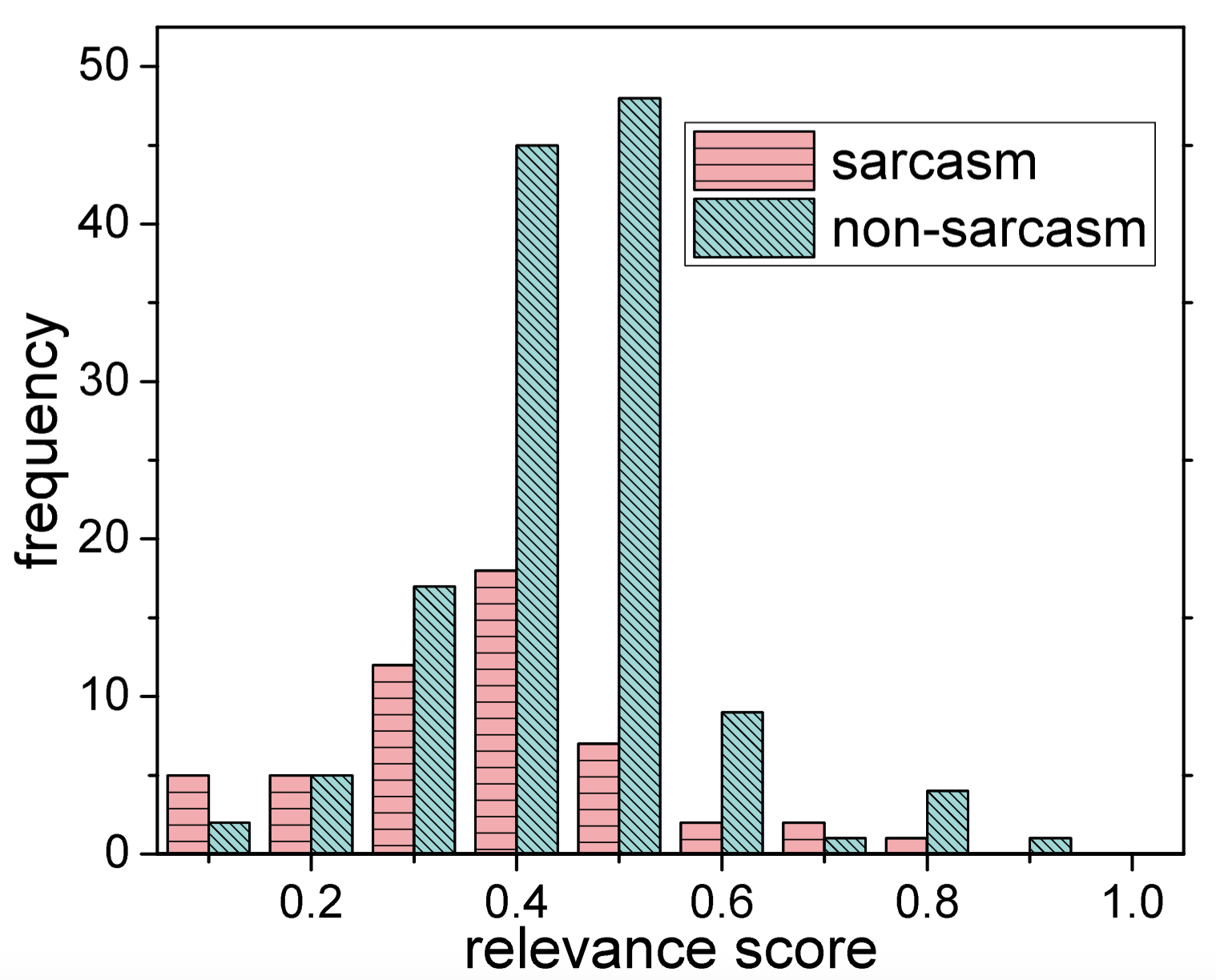}}
\centerline{\small{(a) word ``good''}}
\label{fig:sarcasm_good}
\end{minipage}
\begin{minipage}{0.32\textwidth}
\centerline{\includegraphics[width=\linewidth]{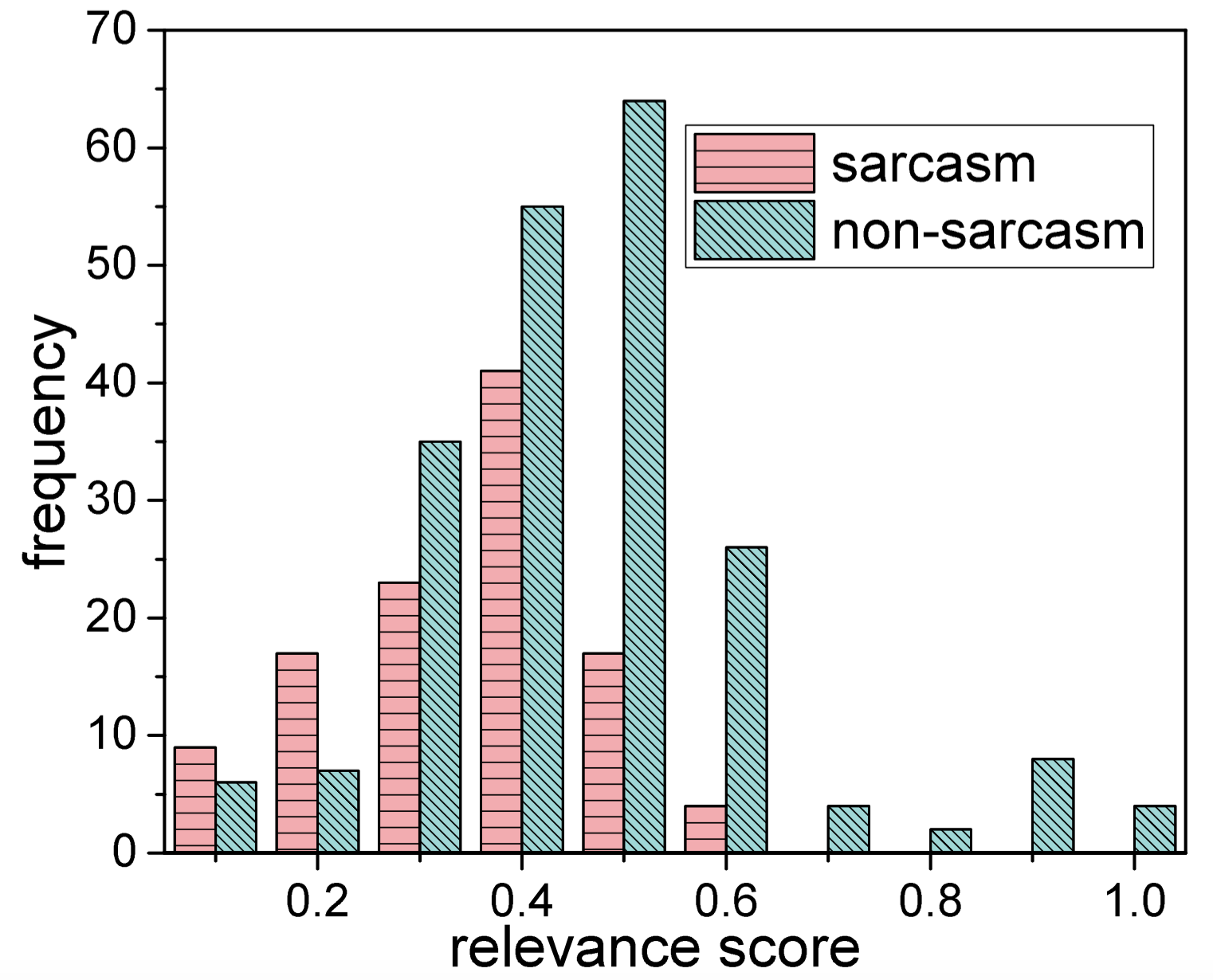}} 
\centerline{\small{(b) word ``love''}}
\label{fig:sarcasm_love}
\end{minipage}
\begin{minipage}[c]{0.32\textwidth}
\centerline{\includegraphics[width=\linewidth]{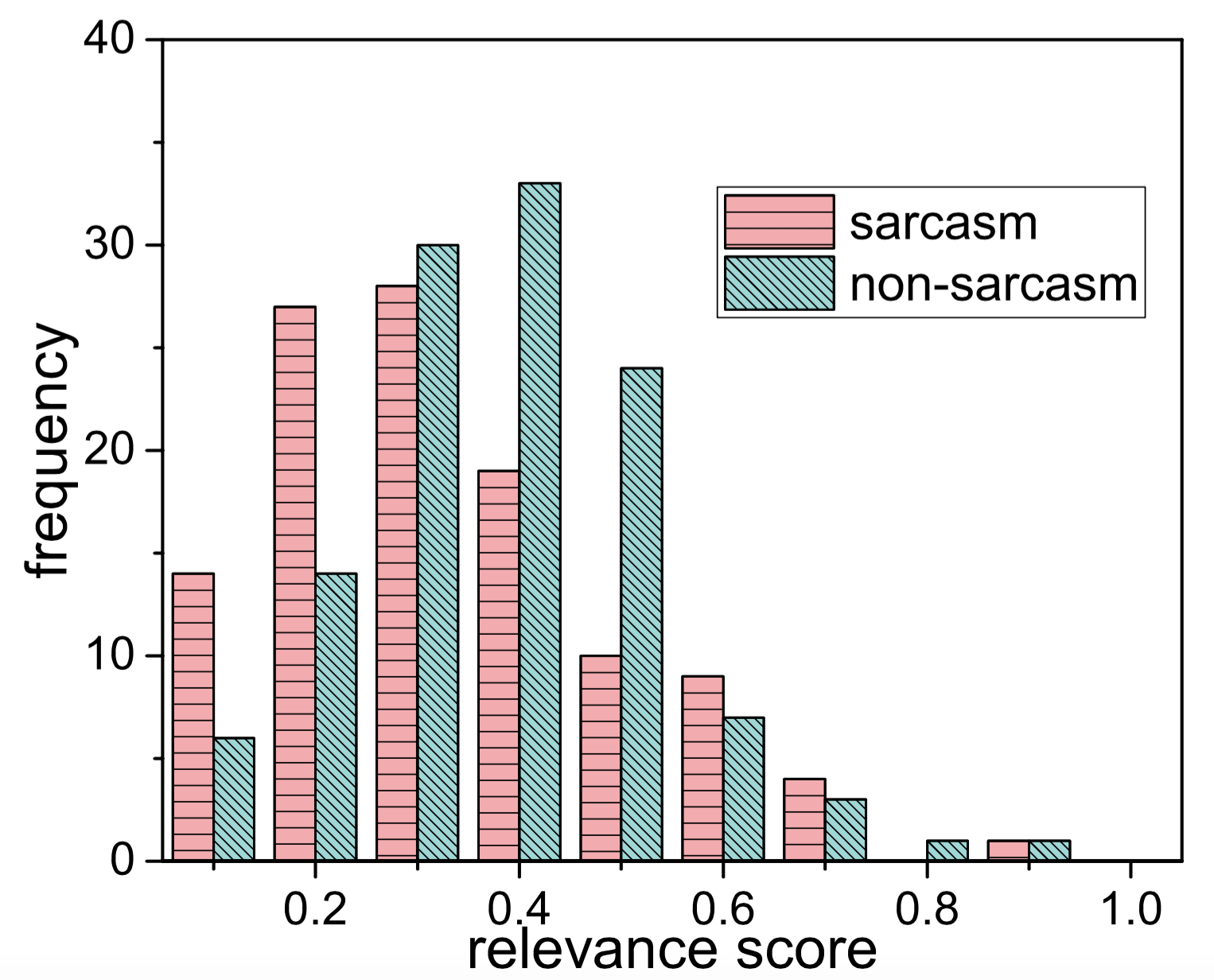}}
\centerline{\small{(c) word ``yeah''}}
\label{fig:sarcasm_yeah}
\end{minipage}
\begin{minipage}[c]{0.32\textwidth}
\centerline{\includegraphics[width=\linewidth]{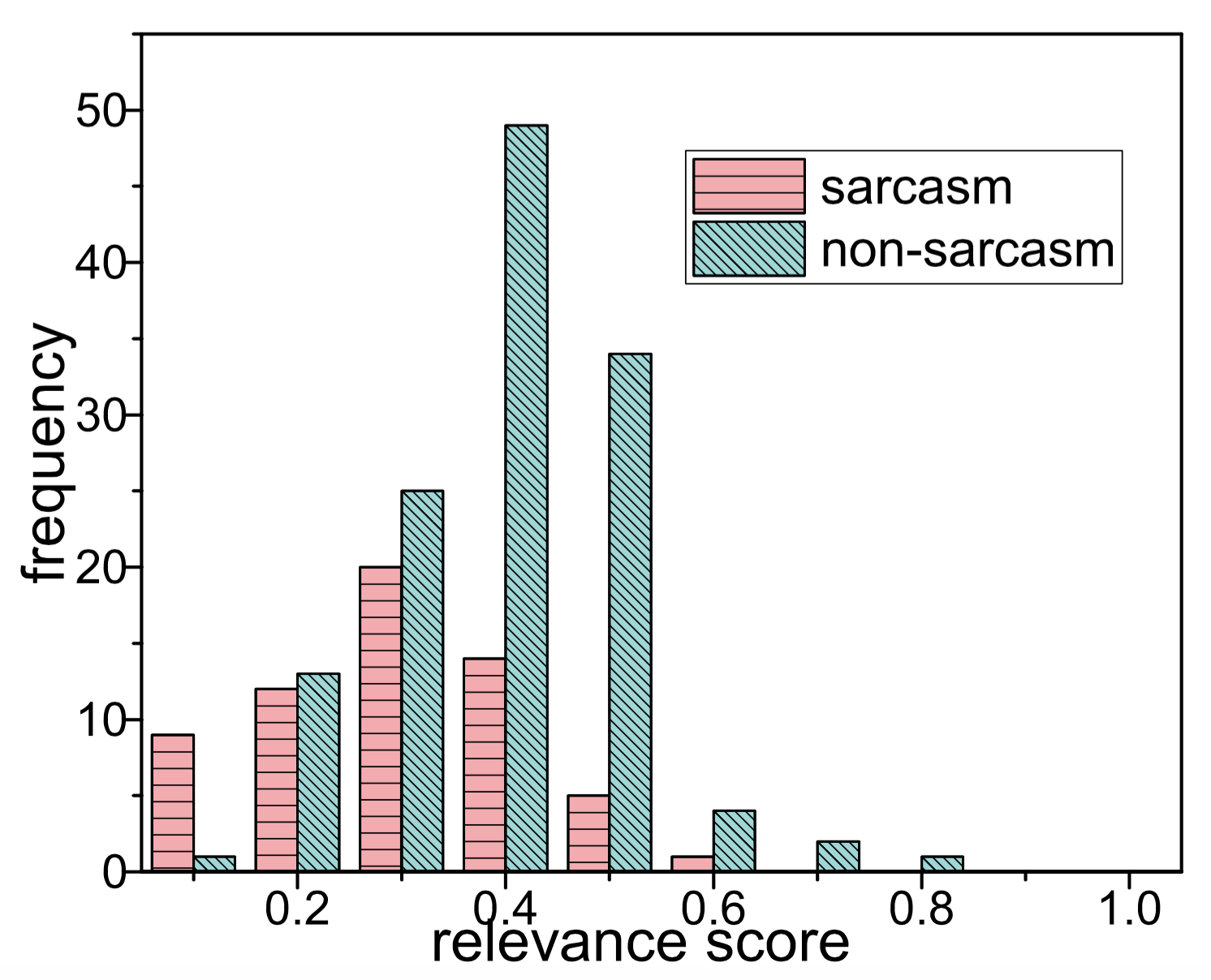}}
\centerline{\small{(d) word ``nice''}}
\label{fig:sarcasm_nice}
\end{minipage}
\begin{minipage}{0.32\textwidth}
\centerline{\includegraphics[width=\linewidth]{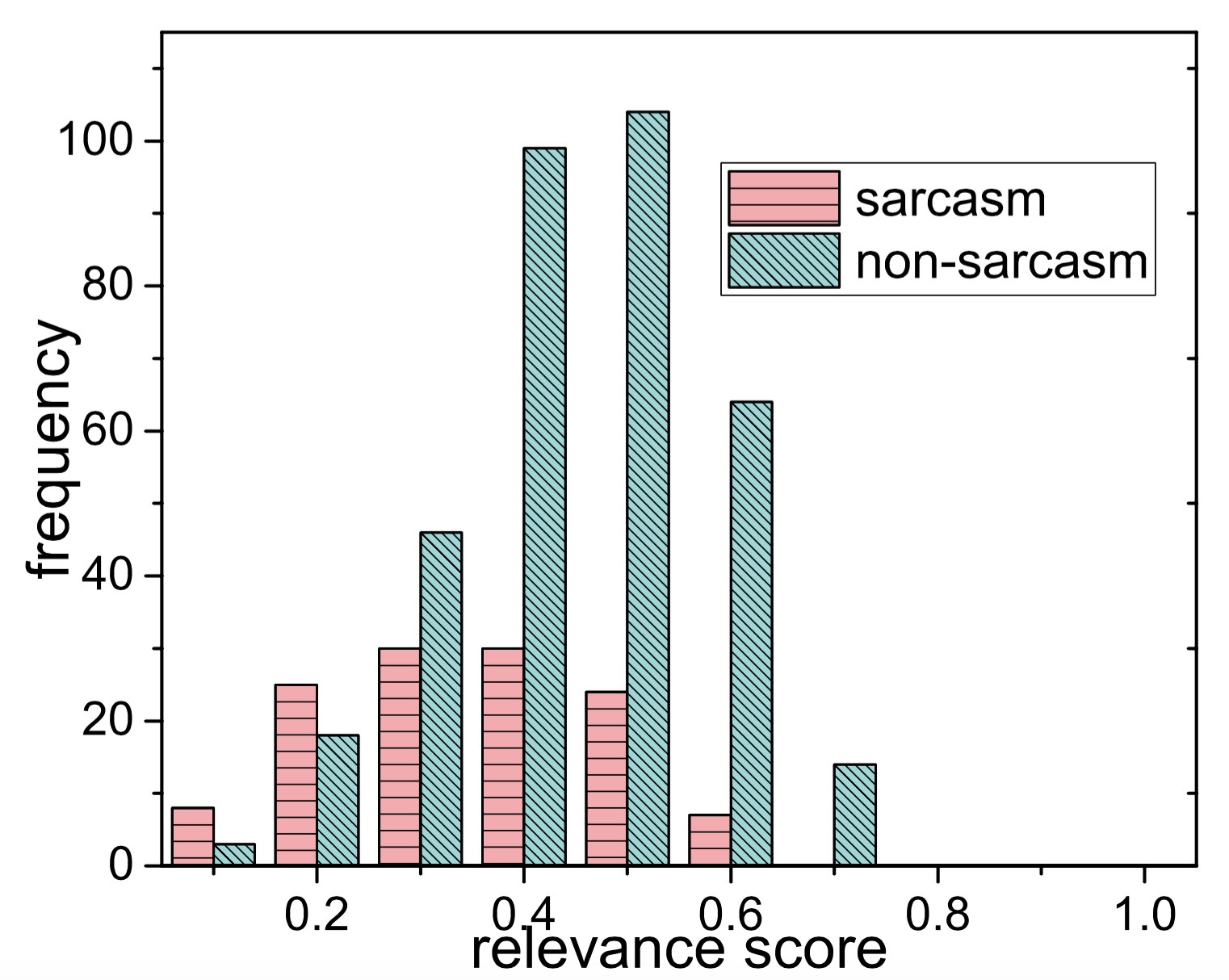}} 
\centerline{\small{(e) word ``always''}}
\label{fig:sarcasm_always}
\end{minipage}
\begin{minipage}{0.32\textwidth}
\centerline{\includegraphics[width=\linewidth]{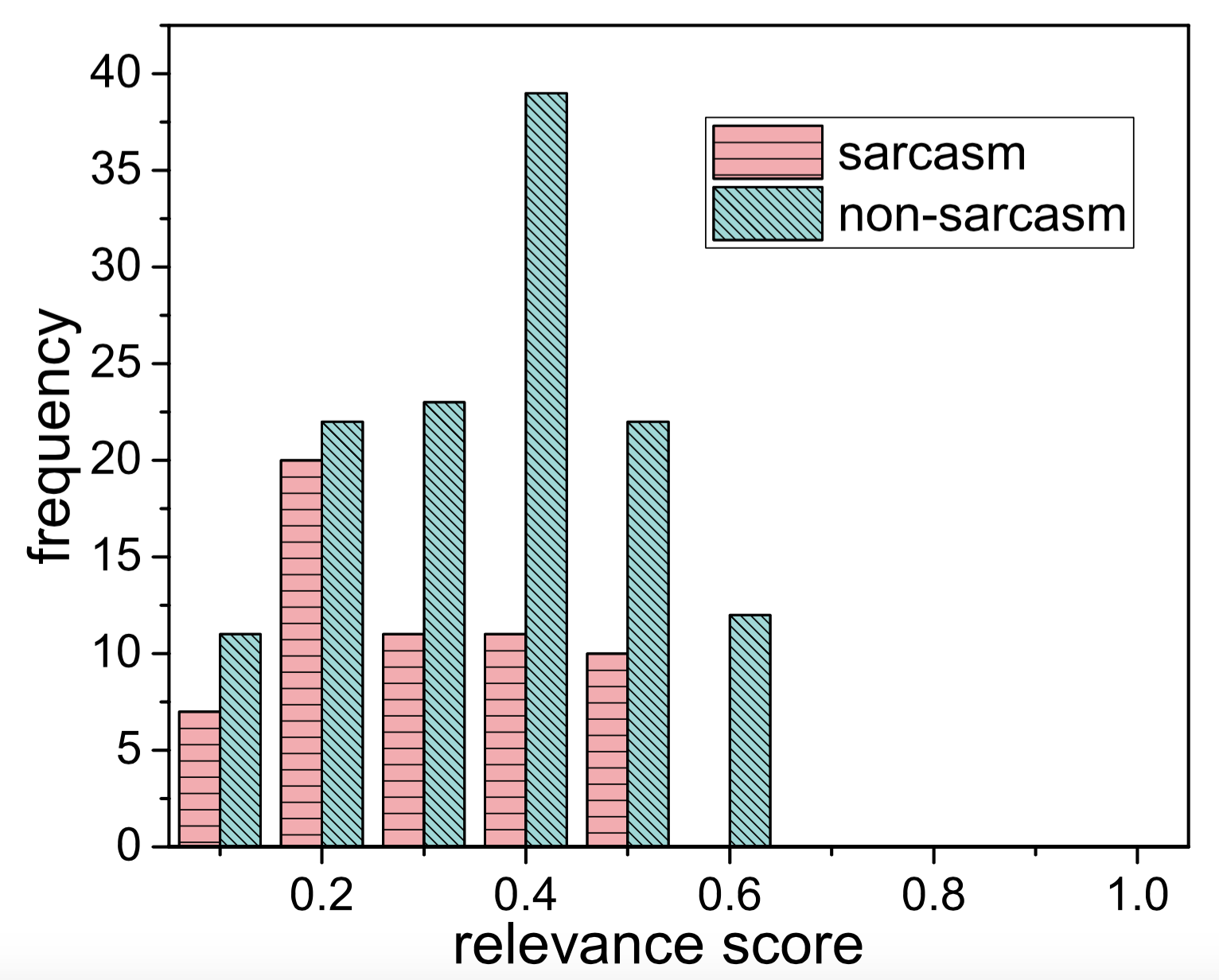}}
\centerline{\small{(f) word ``glad''}}
\label{fig:sarcasm_glad}
\end{minipage}
\caption{Sarcasm Detection in Tweets}
\label{fig:sarcasm}
\end{figure*}

\textbf{Dataset}:
The English noun compounds dataset (ENC), has 90 English noun compounds  annotated on a continuous $[0,5]$ scale for the phrase and component-wise compositionality \cite{reddy2011empirical}; the English verb particle constructions (EVPC) contains 160 English verb-particle compounds, whose componentwise compositionality are annotated on a binary scale \cite{bannard2006acquiring}. German noun compounds (GNC), which contains 246 German noun compounds  annotated on a continuous [1,7] scale for phrase and component compositionality \cite{im2013exploring}. In this paper, we cast compositionality prediction as a binary classification task. We set the same threshold  of 2.5 to ENC as in \cite{salehi2014detecting}, a threshold of 4 to GNC  and use the binary labels of EVPC. The components with score higher than the threshold are regarded as literal, otherwise, they are idiomatic.

{\bf Detection Results}: 
Our subspace representation   (SubSpace) uses CBOW and MSSG embeddings, and we use both average and PCA approximations as context embeddings. Their performance is shown in the row of ``SubSpace (CBOW)'' and ``SubSpace (MSSG)'' respectively.
We have two baseline methods: (1)  pointwise mutual information (PMI). $\text{PMI} = \log\frac{P(w_{1}w_{2})}{P(w_{1})P(w_{2})}$, where $P(\cdot)$ is the probability of the unigram or bigram \cite{manning1999collocations}. %PMI statistically evaluates the cohesion between words. 
Higher PMI indicates the phrase is more likely to be non-compositional. 
%We calculate PMI for each phrase based on English wiki corpus\footnote{available at: \url{https://sites.google.com/site/rmyeid/projects/polyglot} \cite{polyglot:2013:ACL-CoNLL}}, and decide its component words to be non-compositional when its PMI is higher than a threshold. Otherwise, we classify them as compositional. The threshold is trained from training samples.
(2) Average sentence embedding method. While we use PCA, 
%to generate several vectors for sentence semantic representation, 
several recent works have shown  average word vectors to be robust sentence embeddings  \cite{ettinger2016probing,DBLP:journals/corr/AdiKBLG16,wieting2015towards} and we measure compositionality by the cosine similarity between the  target word vector and the sentence vector. The corresponding performance is reported in the rows of ``Avg Cxt (CBOW)'' and ``Avg Cxt (MSSG)''. We only report the best performance of each method in Table \ref{table:mwe-condensed}. \\
%\textbf{and more detailed results are available in supplementary material. \\
We compare with the state-of-the-art of \cite{salehi2014detecting}, specifically their    methods based on word definitions, synonyms and idiom tags (denoted by ALLDEFS+SYN, ITAG+SYN, ALLDEFS) provided by wikitionary. 
%We also report our results on GNC dataset in Table \ref{tab:enc} while we only compare our method with PMI baseline, since Salehi \em{et al.} do not perform experiments on German \cite{salehi2014detecting}. 
As we can see from Table \ref{tab:enc}, our method compares  favorably to the state-of-art performance while outperforming two baseline methods. The key advantage of our method is its non-reliance on external resources like wikitionary or multilingual translations -- these  are heavily relied upon in the state-of-the-art methods \cite{salehi2014detecting,salehi2014using}. Also, unlike the assumption in \cite{salehi2015word}, we {\em do not require that the test phrases appear in the  training corpus}.

\section{Sarcasm Detection}
Sarcasms, also called irony, are  expressions whose actual meaning is quite different - and often opposite to - their literal meaning -- and are instances of non-compositional usage \cite{davidov2010semi,riloff2013sarcasm}. For example, the word `nice' is used in  a sarcastic sense in  `It's so nice that a cute video of saving an animal can quickly turn the comments into politcal debates and racist attacks'. The context  clues identify sarcasm; in this example, `nice' is inconsistent with its context words `debate' and `attacks'. These ideas are used in prior works to create elaborate features (designed based on a large labeled training set) and build a sarcasm detection system \cite{ghoshsarcastic}. We evaluate our compositionality detection algorithm directly on this task. 

\subsection{Qualitative Test}
 {\bf Datasets}: 
 Tweets are ideal sources  of  sarcasm datasets. We use a subset of the tweets in the dataset of  \cite{ghoshsarcastic} and study  words that are used both literally and sarcastically (eg., love, like, favorite, always) for their compositionality.
%In this {\em qualitative} test -- the histogram of compositionality scores for both literal and sarcastic senses are plotted in the supplementary material -- we see that  our algorithm has sufficient power to detect sarcasm in a wide range of contexts. 

We choose six words ``good'', ``love'',  ``yeah'', ``glad'', ``nice'' and ``always'', which   occur with enough frequency in both literal and sarcastic senses in our downloaded dataset. Take the word `nice' as an example and consider its occurrence in the sentence ``It's so nice that a cute video of saving an animal can quickly turn the comments into politcal debates and racist attacks''.  Our compositionality scoring algorithm (cf.\ Section \ref{sec:model}) is applicable here directly: we extract the neighboring content words:\{cute, video, saving, animal, quickly, turn comments, into, political, debates, racist, attacks\} and the result of  PCA on the vectors associated with these words yields the subspace  sentence representation. By projecting the word embedding of `nice' onto this subspace, we get the compositionality score indicating how literal `nice' is in the given sentence. The lower the score, the more sarcastic the word.

%We divide the score range $[0,1]$ into $10$ equal-length segments: $[0, 0.1), [0.1, 0.2),\ldots$. After calculating the scores of the words in given sentences, we count their occurrences in each score segment to see how the scores are distributed for literal and sarcastic word usages.

The histograms of the compositionality scores for these six words ``good'', ``love'',  ``yeah'', ``glad'', ``nice'' and ``always'' (for sarcastic and literal usages) are plotted in Fig.~\ref{fig:sarcasm}. We can visually see that the two histograms (one for sarcastic usage and the other for literal usage) can be distinguished from each other, for each of these three words.  
The histogram of sarcastic usage occupies the low-score region with peak in the $[0.3,0.4)$ bin, while the histogram of literal usage  occupies the high-score region with peak in the $[0.4, 0.5)$ bin.
This  shows that our simple resource-independent compositionality scoring method can distinguish sarcasm and non-sarcasm to some extent.

\begin{table}[]
\centering
\resizebox{0.49\textwidth}{!}{
\begin{tabular}{|l|l|l|l|l|l|l|}
\hline
word     & `good' & `love' & `yeah' & `nice' & `always' & `glad' \\ \hline
accuracy & 0.744  & 0.700  & 0.614  & 0.763  & 0.792    & 0.695  \\ \hline
F1 score & 0.610  & 0.64   & 0.655  & 0.623  & 0.605    & 0.582  \\ \hline
\end{tabular}}
\caption{Twitter Sarcasm Detection}
\label{table:accuracy-sarcasm}
\end{table}

To quantify this extent, we report the accuracy and F1 scores of a simple threshold classifier in each of the six instances in  Table~\ref{table:accuracy-sarcasm}. We emphasize that this performance is derived for a very small dataset (for each of the words) and is achieved using entirely only a trained set of word vectors -- this would be a baseline to build on for the more sophisticated supervised learning systems. 

\subsection{Quantitative Test}
% below is quantitative dataset and test results
\begin{table}[]
\centering
\resizebox{0.47\textwidth}{!}{
\begin{tabular}{|c|c|c|c|c|c|}
\hline
 & \small{Baseline} & \begin{tabular}[c]{@{}c@{}}SubSpace \\(JJ)\end{tabular} & \begin{tabular}[c]{@{}c@{}}SubSpace \\(VB)\end{tabular} & \begin{tabular}[c]{@{}c@{}}SubSpace \\(JJ+ RB)\end{tabular} & \begin{tabular}[c]{@{}c@{}}SubSpace (JJ+\\ RB+VB)\end{tabular} \\ \hline
features & \textgreater50,000 & 2 & 2 & 3 & 4 \\ \hline
precision & 0.315 & 0.278 & 0.289 & 0.279 & 0.278 \\ \hline
recall & 0.496 & 0.936 & 0.844 & 0.98 & 0.747 \\ \hline
F1 score & 0.383 & 0.426 & 0.396 & \textbf{0.434} & 0.393 \\ \hline
\end{tabular}}
\caption{Sarcasm Detection on Reddit Dataset}
\label{tab:sarcasm}
\end{table}
A {\em quantitative} test is provided via our study on 
 a Reddit irony dataset \cite{wallace2014humans}. This dataset consists of 3020 annotated comments containing 10401 sentences in total, and each comment is labeled with ``ironic'', ``don't know'' and ``unironic''. An example of an ironic comment is ``It's amazing how Democrats view money. It has to come from somewhere you idiots and you signed up to foot the bill. Congratulations.'' %An example of an unironic comment is ``You know we can hear you, right? Yes. (continues rant)''. 
  The task is to identify whether a given comment is ironic or not. \cite{wallace2014humans} considers the 50,000 most frequently occurring unigrams and bigrams, and use binary bag-of-words and punctuations as features followed by a linear kernel SVM, grid-search for parameter tuning and five-fold cross validation. Instead, we generate compositionality-score  features based on POS tags to allow a direct  comparison with the state-of-the-art in \cite{wallace2014humans}.   

{\bf Algorithm Description}: 
For a given comment, we first select words that might have sarcastic meaning based on their POS tags: adjectives (like `favorite'), adverbs (like `happily') and verbs (like `love') are likely to be used in irony, and we pick candidate words whose POS tag are JJ (adjective), RB (adverb), or VB (verb). \\
For each of the selected words in a given sentence, we obtain its compositionality score  with respect to its local context. Among all these scores, we choose $k$ smallest scores as features (here $k=2,3,4$  refers to a very small number of features,  cf.\ Table  ~\ref{tab:sarcasm}). 
These features are then fed  into the same supervised learning system as in \cite{wallace2014humans}, providing a fair comparison between the two feature-selection  methods. 

{\bf Results}: 
%The baseline system is provided by  \cite{wallace2014humans}, which involved a large number of features of the context. 
The experiment results of using compositionality scores as features instead are shown in Table \ref{tab:sarcasm} -- we use {\em much fewer} features than the baseline and also get comparable results; indeed in some instances (such as JJ+RB class), we achieve a $5\%$ higher F1 score over the baseline system.
% above is quantitative dataset and test results

% metaphor
\section{Metaphor Detection}
\begin{table}[]
\centering
\resizebox{0.46\textwidth}{!}{
\begin{tabular}{|c|c|c|c|c|}
\hline
 &  & features & accuracy & f1 score \\ \hline
\multirow{3}{*}{\begin{tabular}[c]{@{}c@{}}SVO\end{tabular}} & state-of-art & 279 & {\bf 0.82} & {\bf 0.86} \\ \cline{2-5} 
 & \begin{tabular}[c]{@{}c@{}}SubSpace\\ original sentence\end{tabular} & 4 & 0.729 & 0.744 \\ \cline{2-5} 
 & \begin{tabular}[c]{@{}c@{}}SubSpace\\ longer sentence\end{tabular} & 4 & 0.809 & 0.806 \\ \hline
\multirow{3}{*}{\begin{tabular}[c]{@{}c@{}}AN\end{tabular}} & state-of-art & 360 & {\bf 0.86} & {\bf 0.85} \\ \cline{2-5} 
 & \begin{tabular}[c]{@{}c@{}}SubSpace\\ original sentence\end{tabular} & 3 & 0.735 & 0.744 \\ \cline{2-5} 
 & \begin{tabular}[c]{@{}c@{}}SubSpace\\ longer sentence\end{tabular} & 3 & 0.80 & 0.798 \\ \hline
\end{tabular}}
\caption{Metaphor Detection}
\label{tab:metaphor}
\end{table}
Metaphors are  usually used to express the conceptual sense of a word in noncompositional contexts:  in the sentence ``Comprehensive solutions marry ideas favored by one party and opposed by the other'', the intended meaning of ``marry" is ``combine", a significant (and figurative) generalization of its literal meaning. As such, metaphors form a key part of noncompositional semantics and are natural targets to study in our generic framework. 

{\bf Dataset}: 
English datasets comprising of metaphoric and literal uses of two syntactic structures (subject-verb-object (SVO) and adjective-noun (AN) compounds) are provided in \cite{tsvetkov2014metaphor}. An example of an SVO metaphor is ``The twentieth century saw intensive development of new technologies'', and an example of an AN metaphor is ``black humor seems very Irish to me''. Our task is to decide whether a given sentence containing either SVO or AN structure is used as a metaphor. The SVO dataset contains 111 literal and 111 metaphorical phrases while the AN dataset contains 100 literal and 100 metaphorical phrases.

{\bf Algorithm Description}: 
The state-of-the-art work \cite{tsvetkov2014metaphor} uses  training-data-driven feature engineering methods while relying on external resources like WordNet and the MRC psycholinguistic database. We depart by using the {\em unsupervised} scores generated by our compositionality detection algorithm, albeit specific to POS tags (critical for this particular dataset since it is focused on specific syntactic structures), as features for metaphor detection.

%Recall that our compositionality detection algorithm returns a score of a word given its sentence, and the score reflects how `compositional' the word is in the given context.
For each word in the SVO or AN structure, we obtain a compositionality score with respect to its local context and derive features from these scores: 
The features we derive for the SVO dataset from these scores are: (1) the lowest score in SVO; (2) verb score; (3) ratio between lowest score and highest score; (4) $\min{(\frac{\text{verb score}}{\text{subj score}}, \frac{\text{subj score}}{\text{verb score}}, \frac{\text{verb score}}{\text{obj score}}, \frac{\text{obj score}}{\text{verb score}})}$. 

In the sentence ``The twentieth century saw intensive development of new technologies'', `century' (subject), `saw' (verb) and `development' (object) form the SVO structure. The compositionality scores of the subject, verb and object, are computed as outlined in Section~\ref{sec:model}. 

If an SVO phrase is a metaphor, then we expect there will be at least one word which is inconsistent with the context. Thus we include the lowest score as one of the features. Also, the verb score is a feature since  verbs are frequently used metaphorically in a phrase.  
The absolute score is very sensitive to the context, and we also include relative scores to make the features more robust. The relative scores are the ratio between the lowest score and the highest score, and the minimum ratio between verb and subject or object. 

The features we get for AN dataset from these scores are: (1) the lowest score in AN; (2) the highest score; (3) ratio between the lowest and the highest score. In the sentence ``black humor seems very Irish to me'', `black' (adjective) and `humor' form the AN structure. We calculate compositionality scores for these two words `black' and `humor', and use them to generate the features described above.

These features are then fed into a supervised learning system (random forest), analogous to the one in \cite{tsvetkov2014metaphor} allowing for a fair comparison of the power of the features extracted. 
%The precise  training and testing procedures are detailed in supplement. 

{\bf Detection Results}:
% SVO and AN dataset
%It is worth mentioning that when we get the relevance of a verb, we consider all forms of this verb (past tense, present tense, etc.), and choose the highest relevance score among scores of all forms as the verb score. Such processing gives us more accurate result since different forms of the verb usually have the same semantic meaning. \\
The experimental results on SVO and AN datasets are detailed in Table \ref{tab:metaphor} where the baseline is provided by the results of \cite{tsvetkov2014metaphor} (which has access to the MRC psycholinguistic database and the supersense corpus). On the full set of  original sentences, the performance of our compositionality detection algorithm (with only four features in stark contrast to the more than 100 used in the state of the art) is not too far from the baseline. 

Upon a closer look, we find that some of the original sentences are too short, e.g.\ ``The bus eventually arrived''. Our context-based method naturally does better with longer sentences and we purified the dataset by replacing  sentences whose non-functional words are fewer than 7 with longer sentences extracted from Google Books. We rerun our experiments and the  performance on the longer sentences is improved, although it is still a bit below the baseline -- again, contrast the very large number of features (extracted using significant external resources) used in the baseline to just 3 or 4  of our approach (extracted in a resource-independent fashion).

\section{Related Works}
\label{sec:RelatedWorks}

\noindent {\bf Average sentence approximation}: 
Using the average of word embeddings to represent the sentence is a simple, yet robust, approach in several settings. For instance, such a representation is successfully used for sentential sentiment prediction  \cite{faruqui2015retrofitting} and in  \cite{kenter2015short} to study text similarity. Average word embeddings are also used  \cite{kenter2016siamese} in conjunction with a neural network architecture to predict the surrounding sentences from the input sentence embeddings. Computational models of sentential semantics have also shown to be robustly handled by average word embeddings 
\cite{yu2014deep,gershman2015phrase,DBLP:journals/corr/AdiKBLG16,wieting2015towards}.  In the compositionality testing experiments of this paper, the average  representation performs reasonably well, although the subspace representation is  statistically significantly superior. 

\noindent {\bf Compositionality Detection}: 
Among the  approaches to predict the idomaticity of MWEs, external linguistic resources are  natural sources to rely on.  Early approaches  relied on the use of specific lexical and syntactic properties of MWEs \cite{lin1999automatic,mccarthy2003detecting,cook2007pulling,fazly2007distinguishing}.  More recent approaches include multilingual translations  \cite{salehi2013predicting,salehi2014using} and using  Wikitionary (one approach uses its word definitions, idiom tagging together with word synonyms to classify idiomatic phrases) \cite{salehi2014detecting}. By their very nature, these approaches have limited coverage of semantics and are highly language dependent.  

In terms of distributed representation, methods include Latent Semantic Analysis  \cite{LSA}  and word embeddings which have been extaordinarily successful representations of word semantics,  eg., word2vec and GloVe \cite{mikolov2014word2vec,pennington2014glove,neelakantan2015efficient}. 
\cite{salehi2015word} is a recent work  exploring compositionality in conjunction with word embeddings; however, an aspect not considered  is that compositionality does not only depend on the phrase but also on its context -- this results in an  inability to identify the context-based compositionality of polysemous phrases like \emph{bad egg}. 

\noindent{\bf  Sarcasm Detection} 
Sarcasm is a figurative expression conveying a meaning that is  opposite of its literal one, usually  in an implicit way, and is a crucial component in {\em sentiment analysis}. Such connections are explored in \cite{maynard2014cares} via a rule-based method of identifying known sarcastic phrases. %
Semi-supervised sarcasm identification algorithms are identified in \cite{davidov2010semi,riloff2013sarcasm,liebrecht2013perfect,maynard2014cares}, each using different sets of features (eg., word senses, uni, bi and trigrams) that are then fed into a classification system tuned on a large training dataset.   %\cite{riloff2013sarcasm}  distinguishes the phrase usages in positive and negative situations as related to sarcasms, which is then identified based on the similarity between a test example and each situation. \cite{liebrecht2013perfect} uses uni-, bi- and trigrams as features into a supervised (Winnow) classifier system of sarcasm detection 
%\cite{maynard2014cares}. \cite{ghoshsarcastic} consider  sarcasm detection to be word sense disambiguation task, and train word embeddings with a large labeled Twitter dataset. %The disambiguation decision is based on the similarity of the test context and the trained literal and sarcastic contexts. 

\noindent {\bf Metaphor Detection} 
% http://www.aclweb.org/anthology/P14-1024
Metaphors offer figurative interpretations and are a key feature of natural language  \cite{lakoff1980conceptual}.
%\textbf{Linguistic resource-based method} 
\cite{mason2004cormet} considers  metaphor expression as a mapping from a source domain to a target domain, and  develops a corpus-based system, CorMet, to discover such metaphorical equivalances  based on WordNet.
\cite{turney2011literal} hypothesises that metaphorical usage is related to the degreee of contextual abstractness, which they quantify relying on the  MRC Psycholinguistic Database Machine Usable Dictionary (MRCPD) \cite{coltheart1981mrc}. 

\cite{broadwell2013using} proposes  a detection method according to  lexical imaginability, topic chaining and semantic clustering. Their method is also based on the  linguistic resource of MRCPD.
\cite{tsvetkov2014metaphor} focuses on Subject-Verb-Object and Adjective-Noun structures, and use word abstractness and imagineability as well as supersenses as features for metaphor detection. Besides MRCPD, they also have recourse to WordNet for word supersenses.

\section{Conclusion}
\label{sec:conclusion}
We bring  MWEs, sarcasms and metaphors under a common umbrella of compositionality,  followed by a simple unified framework to study it; this is our central contribution. The method proposed to detect word/phrase compositionality based on local context is simple and  affords  a clear geometric view. We do not depend on  external resources and perform very well across multiple languages and in a large variety of settings (metaphors, sarcastic and idiomatic usages). The method naturally scales to handle complications such as unseen  phrases and polysemy, achieving comparable or superior results to the state-of-art (which are supervised methods based on elaborate feature engineering and using, at times, plentiful external linguistic resources) on standard datasets.

A careful understanding of the geometry of our  context representations (subspace of the principle components of the word vectors) and compositionality scoring method, along with  a study of the connections to neural network methods of sentence representation (eg., LSTM \cite{greff2015lstm}) are  interesting future avenues of research.

\newpage
\bibliography{tacl}
\bibliographystyle{acl2012}

\end{document}